\documentclass{article} 
\usepackage{colm2024_conference}

\usepackage{microtype}
\usepackage{hyperref}
\usepackage{url}
\usepackage{booktabs}
\definecolor{darkblue}{rgb}{0, 0, 0.5}
\hypersetup{colorlinks=true, citecolor=darkblue, linkcolor=darkblue, urlcolor=darkblue}

\usepackage{graphicx}
\usepackage{textcomp}
\usepackage{adjustbox}
\usepackage{multirow}
\usepackage{amsmath}
\usepackage{booktabs}
\usepackage{caption}
\usepackage{subcaption}
\usepackage{hyperref}

\title{Effectively Prompting Small-sized Language Models for Cross-lingual Tasks via Winning Tickets}

\begin{document}

\maketitle

\begin{abstract}
Current soft prompt methods yield limited performance when applied to small-sized models (fewer than a billion parameters). Deep prompt-tuning, which entails prepending parameters in each layer for enhanced efficacy, presents a solution for prompting small-sized models, albeit requiring carefully designed implementation. In this paper, we introduce the Lottery Ticket Prompt-learning (LTP) framework that integrates winning tickets with soft prompts. The LTP offers a simpler implementation and requires only a one-time execution. We demonstrate LTP on cross-lingual tasks, where prior works rely on external tools like human-designed multilingual templates and bilingual dictionaries, which may not be feasible in a low-resource regime. Specifically, we select a subset of parameters that have been changed the most during the fine-tuning with the Masked Language Modeling objective. Then, we prepend soft prompts to the original pre-trained language model and only update the selected parameters together with prompt-related parameters when adapting to the downstream tasks. We verify the effectiveness of our LTP framework on cross-lingual tasks, specifically targeting low-resource languages. Our approach outperforms the baselines by only updating 20\% of the original parameters.
\end{abstract}

\section{Introduction}

Current research has empirically validated that prompting approaches, which involve reformulating downstream tasks as language modeling problems, align well with pre-training objectives and thus yield improved performance (\citealp{NEURIPS2020_1457c0d6}; \citealp{le-scao-rush-2021-many}). Prompt-tuning has shown improvements comparable to fine-tuning, distinguishing itself among various prompting methods. The key idea is to prepend a sequence of continuous tokens and tune them while keeping the pre-trained model frozen. However, its performance tends to be less optimal than fine-tuning for models with fewer than 10 billion parameters (\citealp{lester-etal-2021-power}). This discrepancy in performance is even more pronounced for models with only millions of parameters. P-tuning v2 (\citealp{liu-etal-2022-p}) provided a solution for small-sized models while requiring a meticulously designed implementation of Deep Prompt Tuning (\citealp{li-liang-2021-prefix}; \citealp{qin-eisner-2021-learning}). Another strategy for adapting small-sized models to specific tasks is Prompt+LM tuning, which fine-tunes both the prompt-relevant parameters and all the parameters of the pre-trained model simultaneously. However, this fine-tuning process bears the risk of overfitting when engaging with extremely small datasets due to the vast number of parameters involved.

Upon employing Prompt+LM tuning to adapt multilingual language models to low-resource languages, its constraints become more pronounced. The key challenge in these contexts is data scarcity, which significantly undermines the suitability of the Prompt+LM method. Moreover, the extensive tuning of parameters for a single language could inadvertently compromise the model's performance across other languages. Some research efforts (\citealp{qi2022enhancing}; \citealp{li-etal-2023-enhancing-cross}) have tackled these challenges by incorporating external expert knowledge, yet these approaches introduce additional complexities and lack flexibility in adapting to other languages.

\begin{figure*}[h]
  \centering
    \includegraphics[width=14cm]{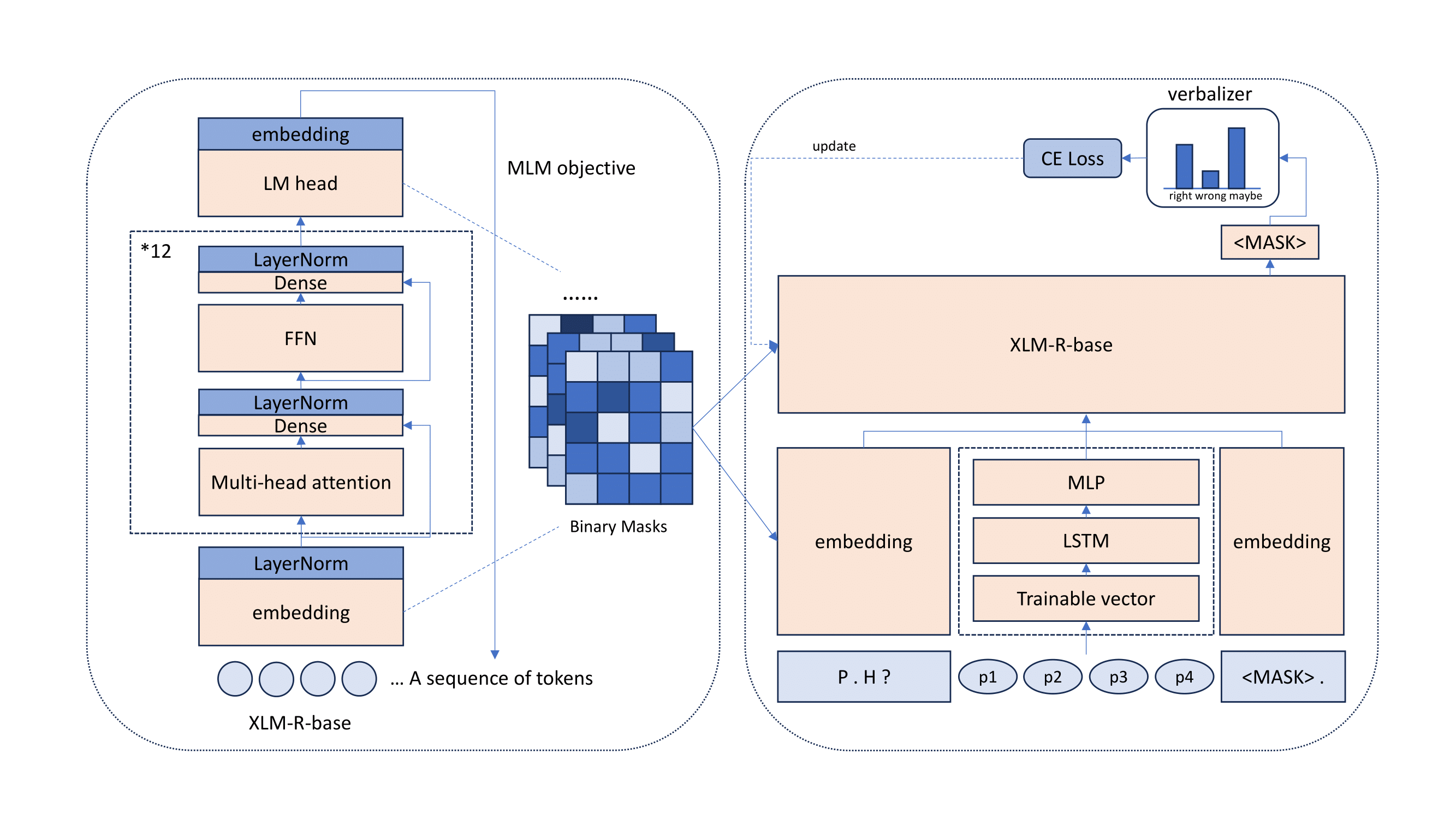}
  \caption{Lottery Ticket Prompt-learning framework: a) the parameter selection step is presented on the left, where frozen parameters are denoted in blue, while trainable parameters are highlighted in orange; b) the prompt-learning step is depicted on the right, where the generated binary masks keep certain parameters unchanged.}\label{figure:model}
\end{figure*}

To address these problems, we propose Lottery Ticket Prompt-learning (LTP), a simple yet efficient framework that could effectively prompt small-sized language models. Inspired by the Lottery Ticket Hypothesis (\citealp{frankle2018the}), our approach selectively prompts a subset of the model's parameters, ensuring that the tuning process does not significantly disrupt the model's language-specific knowledge. This approach enables us to blend the advantages of winning tickets with prompting, leading to enhanced performance. Specifically, we select the subset of parameters that are more active than others with the Masked Language Modeling objective on the English Wikipedia dataset. Then, we prepend a sequence of continuous vectors to the input and initialize the language model using the original pre-trained parameters. We conduct prompt-learning only on the prompt-related parameters and the selected subset of parameters. Our method is illustrated in Figure~\ref{figure:model}.

Our main contribution in this paper is a new solution for prompting small-sized language models. We offer a more straightforward implementation that doesn't necessitate meticulous design and only requires a one-time execution. Moreover, our approach facilitates adaptation to low-resource languages, both seen and unseen by the pre-trained models, by reducing tuned parameter sizes without significantly altering language-specific knowledge. It also eliminates the reliance on external tools like dictionaries or translation models, which are often unfeasible for low-resource language contexts.

We demonstrate the effectiveness of our method on XNLI and AmericasNLI datasets. Experimental results show that our LTP framework outperforms baselines using only 20\% of the original parameters. Additionally, we find that the active parameters in the middle layers are more expressive and thus could further shrink the size of trainable parameters.


\section{Background}

\subsection{Prompting}

Starting from GPT-3 (\citealp{NEURIPS2020_1457c0d6}), prompt-related methods have emerged as a promising means for exploring the internal knowledge of language models. They incorporate the input into a template, either human-designed or auto-generated, to reformulate the original tasks as language modeling tasks aligned with the pre-training procedure. For example, we have a movie review, "Emotionally Charged and Brilliantly Crafted" to classify. We could add a template "This movie was \_" after the review and generate the answer probability of the next token using the language model. Then, we use a verbalizer to map words into different categories ("great" \textrightarrow "positive"; "terrible" \textrightarrow "negative"). This paradigm leverages in-context learning without any steps to update parameters. However, it is not applicable to small-sized pre-trained models (fewer than 1 billion parameters), as the model scale constrains its effectiveness.

Exploring prompts for small-sized models involves three distinct lines of research: Fixed-LM Prompt Tuning, Fixed-prompt LM Tuning, and Prompt+LM Tuning. Here, we use the definition of categorization according to \citet{liu2023pre}.

\textbf{Fixed-LM Prompt Tuning} only updates prompt-related parameters and keeps the backbone model unchanged. This setting is parameters-efficient, while its performance is constrained when applied to small-sized models, and, to the best of our knowledge, remains unexplored for Bert-base sized models (110M parameters).

\textbf{Fixed-prompt LM Tuning} typically employs a discrete prompt like the movie review template we introduced earlier and updates all or some of the parameters in the backbone model. The performance is prone to the prompt in this scenario.

\textbf{Prompt+LM Tuning} refers to updating prompt-related parameters and all or some of the parameters in the backbone model. The prompt contains continuous tokens not from the vocabulary. This setting requires training more parameters than the other two and might result in overfitting in a few-shot scenario, but it comes with corresponding improvements. Both \citet{zhao-schutze-2021-discrete} (Soft prompting and Mixed prompting) and our approach align with this research line, but we require fewer parameters and achieve more gains.

\subsection{Lottery Ticket Hypothesis}

The Lottery Ticket Hypothesis concluded that training a sub-network in isolation can match the performance of training the entire network, and they defined this trainable sub-network as winning tickets. The training procedure starts with identifying a winning ticket, which is achieved by training the network and pruning weights with the smallest magnitudes. Each unpruned weight is then reset to its initial state from the original network before training and re-tuned. Lottery Ticket Hypothesis has been applied for model compression, mitigating the side effects of noisy labels, and adapting pre-trained models. To the best of our knowledge, we are the first to integrate it into cross-lingual prompt-learning.

\section{The LTP framework}

The LTP framework integrates Lottery Ticket Hypothesis and prompt-based fine-tuning, which only tunes prompt-related parameters and a subset of parameters of the language model. Figure~\ref{figure:model} depicts the proposed LTP framework.

\subsection{Lottery ticket fine-tuning}

We select the language-agnostic parameters to integrate into prompt-learning, addressing two major drawbacks of the original prompt+LM Tuning approach. First, training a large amount of parameters could exacerbate potential overfitting issues in few-shot scenarios. Second, fine-tuning the entire model using a few samples from a source language and then transferring (zero-shot) it to others may affect their expressiveness and performance.

Inspired by \citet{ansell-etal-2022-composable}, we could find a sparse network adapted to a specific language. We have a pre-trained language model, denoted as M, with parameters represented by $\theta$. We update all the trainable parameters using the Masked Language Modeling objective on the dataset of a selected language. This study employs English for active parameter detection due to its status as a high-resource language. The parameters associated with English could convey richer semantic and structural information, making them valuable for transfer to other languages. The updated parameters are denoted as $\theta^l$. Then, we rank the parameters based on the greatest absolute difference $|\theta_i^l-\theta_i|$. In this manner, we can select the top K parameters that have changed the most. We set a threshold $\mu$ to filter K active parameters. We store the selected parameters to make this process a one-time event and exclude other parameters from being included in subsequent fine-tuning steps. We apply L1 regularization in the experiments to discourage extensive parameter deviations from their pre-trained values.

In addition, we explore two strategies to address the issue of parameter selection bias towards the embedding matrix: (a) decoupling and freezing the LM head's embedding parameters to prevent downstream performance impact, and (b) freezing the embedding layer to focus parameter updating on the Transformer layers.

\subsection{Prompting and sparse LM fine-tuning}

We adopt Soft Prompt in this stage, as it eliminates the need for external tools (or experts) and seamlessly adapts to various languages. Soft Prompt is a sequence of "pseudo tokens" that are not from the vocabulary, while it empowers the language model to generate the answers using its internal knowledge. Consider a cross-lingual natural language inference task dataset consisting of labeled triplets ${(P_l^i, H_l^i, Y^i)}$, where $P_l^i$ and $H_l^i$ denote a pair of premise and hypothesis in a language $l$, and $Y^i\in\{entailment, contradiction, neutral\}$ denotes the label. We reformulate this task using soft prompts into a Masked Language Modeling task, which is used during the pre-training process:

$P_i$ . $H_i$ ? $<p_1><p_2>...<p_n> <MASK>$.

where $p_i, i\in[1,n]$ is a soft token associated with a randomly initialized trainable vector, the output dimension of which is equivalent to the hidden dimension of the original embedding layer. We reparameterize $p_i$ with a bidirectional LSTM layer and an MLP head to yield better performance. Prompt-related parameters include the trainable vector, the LSTM layer, and the MLP head. Then, we prepend the output of the MLP head (which is considered as soft prompt embeddings) to the embeddings of other tokens and feed this concatenation into Transformer layers.


In the fine-tuning step, we first reset all the parameters of the backbone model to the original values $\theta$. Then, we reformulate the task as above, and we can get the probability distributions of the $<MASK>$ token. We use a verbalizer ("entailment" \textrightarrow "right"; "contradiction" \textrightarrow "wrong"; "neutral" \textrightarrow "maybe") to map the output tokens to labels. We only update the prompt-related parameters and the parameters we selected in the first stage while freezing others. Compared to the approach in \citet{zhao-schutze-2021-discrete}, we diminish the size of trainable parameters in the prompt+LM Tuning paradigm, thereby enhancing both efficiency and effectiveness.

\section{Experiments}

\subsection{Datasets}

\textbf{MLM Training Data.} During the parameters selection process, we only require unlabeled English text. Therefore, we randomly select 300K English sentences from the Wikipedia dataset (\citealp{wikidump}), identified for its comprehensive coverage of general knowledge.

\textbf{XNLI.} Cross-lingual natural language inference requires a comprehensive understanding of contextual complexities. XNLI (\citealp{conneau-etal-2018-xnli}) expands SNLI/MultiNLI (\citealp{bowman-etal-2015-large}; \citealp{williams-etal-2018-broad}) across fifteen languages. Our proposed LTP framework is designed to focus on the few-shot scenario, so we use training and development data sampled by \citet{zhao-schutze-2021-discrete} for a fair comparison. In particular, we have $K \in \{1,2,4,8,16,32\}$ shots per class from the MNLI training set and the corresponding translations in other languages used for the in-language prompting setting. This allows for simulating a realistic low-resource regime, providing a thorough evaluation in scenarios with limited data availability.

\textbf{AmericasNLI.} Truly low-resource languages that remain unexposed to commonly pre-trained language models present a more significant challenge. AmericasNLI (\citealp{ebrahimi-etal-2022-americasnli}) is an extension of XNLI, incorporating 10 Indigenous languages prevalent in the Americas. We randomly selected 32 samples per class from AmericasNLI's development set for both training and development sets.

\begin{table*}[!ht]
\centering
\begin{adjustbox}{max width=0.95\textwidth}
\begin{tabular}{c|c|ccccccccccccccc|c}
\hline
\textbf{Shots} & \textbf{Method} & \textbf{AR} & \textbf{BG} & \textbf{DE} & \textbf{EL} & \textbf{EN} & \textbf{ES} & \textbf{FR} & \textbf{HI} & \textbf{RU} & \textbf{SW} & \textbf{TH} & \textbf{TR} & \textbf{UR} & \textbf{VI} & \textbf{ZH} & \textbf{AVG.}\\
\hline
\multirow{6}{*}{1} & \text{FT} & \text{32.53} & \text{32.63} & \text{32.94} & \text{32.53} & \text{32.91} & \text{32.61} & \text{32.65} & \text{32.87} & \text{32.67} & \text{32.77} & \text{33.11} & \text{32.68} & \text{32.87} & \text{32.69} & \text{32.77} & \text{32.75} \\
& \text{DP} & \text{32.08} & \text{33.23} & \text{32.97} & \text{33.24} & \text{33.15} & \text{33.78} & \text{34.08} & \text{33.41} & \text{33.78} & \text{33.45} & \text{33.00} & \text{34.01} & \text{31.99} & \text{32.83} & \text{33.64} & \text{33.24} \\
& \text{SP} & \text{34.84} & \text{36.50} & \text{36.87} & \text{37.49} & \text{36.65} & \textbf{38.29} & \textbf{38.57} & \textbf{36.43} & \text{37.56} & \text{34.52} & \text{35.71} & \textbf{34.76} & \textbf{35.54} & \text{35.06} & \textbf{37.61} & \text{36.43} \\
& \text{MP} & \text{32.31} & \text{32.32} & \text{33.03} & \text{32.14} & \text{33.29} & \text{34.02} & \text{33.74} & \text{34.12} & \text{33.03} & \text{32.86} & \text{32.18} & \text{34.59} & \text{32.65} & \text{32.82} & \text{33.35} & \text{33.10} \\
& \text{LTP\_20\%} & \text{38.38} & \text{39.12} & \text{36.85} & \text{37.56} & \text{39.14} & \text{36.23} & \text{36.85} & \text{34.89} & \text{40.30} & \text{37.50} & \text{38.42} & \text{33.65} & \text{34.49} & \text{38.06} & \text{36.79} & \text{\underline{37.22}} \\
& \text{LTP\_75\%} & \textbf{38.62} & \textbf{39.48} & \textbf{37.68} & \textbf{38.06} & \textbf{39.92} & \text{37.90} & \text{38.22} & \text{35.49} & \textbf{40.88} & \textbf{37.76} & \textbf{39.12} & \text{34.39} & \text{34.95} & \textbf{38.82} & \text{37.60} & \textbf{37.93} \\
\hline
\multirow{6}{*}{2} & \text{FT} & \text{33.16} & \text{33.35} & \text{33.82} & \text{33.24} & \text{33.43} & \text{33.31} & \text{33.30} & \text{33.24} & \text{33.29} & \text{33.19} & \text{33.40} & \text{33.04} & \text{33.20} & \text{33.03} & \text{33.29} & \text{33.29} \\
& \text{DP} & \text{32.90} & \text{35.11} & \text{34.44} & \text{34.69} & \text{35.41} & \text{35.43} & \text{34.77} & \text{34.11} & \text{34.93} & \text{32.97} & \text{35.43} & \text{35.19} & \text{32.75} & \text{33.28} & \text{36.46} & \text{34.52} \\
& \text{SP} & \text{35.91} & \text{38.08} & \text{38.15} & \text{38.42} & \text{37.97} & \text{38.23} & \text{38.62} & \text{36.32} & \text{39.22} & \text{34.35} & \text{37.20} & \text{34.75} & \text{35.52} & \text{36.67} & \text{37.71} & \text{37.14} \\
& \text{MP} & \text{32.76} & \text{34.25} & \text{34.10} & \text{33.26} & \text{34.59} & \text{33.81} & \text{34.33} & \text{33.75} & \text{34.01} & \text{33.88} & \text{34.55} & \text{34.51} & \text{32.59} & \text{33.83} & \text{35.39} & \text{33.97} \\
& \text{LTP\_20\%} & \textbf{39.38} & \textbf{39.82} & \text{39.40} & \text{39.90} & \text{40.38} & \text{39.46} & \text{39.84} & \text{38.30} & \textbf{41.26} & \textbf{36.49} & \textbf{39.14} & \text{35.11} & \text{36.99} & \textbf{38.56} & \textbf{38.92} & \textbf{38.86} \\
& \text{LTP\_75\%} & \text{39.00} & \text{38.14} & \textbf{39.88} & \textbf{40.40} & \textbf{41.36} & \textbf{39.50} & \textbf{40.76} & \textbf{38.62} & \text{40.04} & \text{34.95} & \text{38.36} & \textbf{36.03} & \textbf{37.27} & \text{37.60} & \text{37.84} & \text{\underline{38.65}} \\
\hline
\multirow{6}{*}{4} & \text{FT} & \text{33.86} & \text{33.89} & \text{33.73} & \text{33.63} & \text{33.90} & \text{33.58} & \text{33.55} & \text{33.86} & \text{33.58} & \text{33.75} & \text{33.71} & \text{33.79} & \text{33.67} & \text{33.85} & \text{33.78} & \text{33.74} \\
& \text{DP} & \text{35.42} & \text{37.64} & \text{38.85} & \text{37.67} & \text{39.50} & \text{38.91} & \text{38.26} & \text{36.43} & \text{37.54} & \text{34.72} & \text{37.76} & \text{37.23} & \text{35.92} & \text{36.02} & \text{38.74} & \text{37.37} \\
& \text{SP} & \text{38.04} & \text{40.46} & \text{40.08} & \text{40.79} & \text{41.84} & \text{39.78} & \text{41.10} & \text{37.55} & \text{41.72} & \text{35.81} & \text{39.23} & \text{35.88} & \text{37.66} & \text{37.86} & \text{39.48} & \text{39.15} \\
& \text{MP} & \text{33.14} & \text{33.79} & \text{35.16} & \text{33.95} & \text{36.26} & \text{35.52} & \text{35.44} & \text{34.63} & \text{34.21} & \text{33.53} & \text{35.96} & \text{35.62} & \text{33.51} & \text{34.06} & \text{37.10} & \text{34.79} \\
& \text{LTP\_20\%} & \textbf{41.12} & \textbf{41.88} & \text{39.34} & \text{41.46} & \textbf{43.29} & \text{38.94} & \text{40.58} & \text{35.17} & \text{42.83} & \textbf{37.58} & \textbf{41.42} & \text{34.37} & \text{37.37} & \textbf{40.32} & \text{37.98} & \text{\underline{39.58}} \\
& \text{LTP\_75\%} & \text{39.96} & \text{41.84} & \textbf{41.08} & \textbf{41.86} & \text{43.01} & \textbf{40.82} & \textbf{42.73} & \textbf{38.62} & \textbf{42.91} & \text{37.25} & \text{40.82} & \textbf{36.67} & \textbf{39.28} & \text{39.52} & \textbf{40.96} & \textbf{40.49} \\
\hline
\multirow{6}{*}{8} & \text{FT} & \text{32.85} & \text{32.75} & \text{33.05} & \text{32.59} & \text{33.06} & \text{32.58} & \text{32.80} & \text{32.89} & \text{32.88} & \text{32.75} & \text{33.14} & \text{32.69} & \text{33.05} & \text{32.83} & \text{32.65} & \text{32.84} \\
& \text{DP} & \text{32.73} & \text{34.78} & \text{34.79} & \text{34.82} & \text{36.39} & \text{34.97} & \text{35.17} & \text{33.00} & \text{34.59} & \text{32.91} & \text{35.14} & \text{34.13} & \text{33.14} & \text{33.66} & \text{35.56} & \text{34.39} \\
& \text{SP} & \text{36.30} & \text{38.84} & \text{38.22} & \text{38.68} & \text{39.02} & \text{38.16} & \text{38.82} & \text{35.86} & \text{39.73} & \text{34.50} & \text{37.90} & \text{35.11} & \text{35.61} & \text{37.41} & \text{37.17} & \text{37.42} \\
& \text{MP} & \text{32.67} & \text{33.24} & \text{34.81} & \text{33.18} & \text{34.78} & \text{34.66} & \text{34.77} & \text{34.76} & \text{33.81} & \text{33.07} & \text{34.46} & \text{35.12} & \text{32.69} & \text{33.57} & \text{36.34} & \text{34.13} \\
& \text{LTP\_20\%} & \text{37.43} & \text{39.54} & \textbf{39.48} & \text{39.16} & \textbf{40.02} & \textbf{40.22} & \textbf{39.70} & \textbf{38.32} & \textbf{40.62} & \text{35.93} & \textbf{38.66} & \textbf{38.54} & \text{37.43} & \text{37.43} & \textbf{38.58} & \textbf{38.74} \\
& \text{LTP\_75\%} & \textbf{38.44} & \textbf{39.58} & \text{39.06} & \textbf{39.74} & \text{39.74} & \text{39.84} & \text{39.16} & \text{37.23} & \text{40.54} & \textbf{37.07} & \text{38.16} & \text{37.78} & \textbf{37.01} & \textbf{38.22} & \text{37.60} & \text{\underline{38.61}} \\
\hline
\multirow{6}{*}{16} & \text{FT} & \text{33.72} & \text{34.09} & \text{34.28} & \text{33.49} & \text{34.73} & \text{33.82} & \text{33.81} & \text{33.08} & \text{34.06} & \text{33.69} & \text{33.06} & \text{33.57} & \text{33.22} & \text{34.01} & \text{33.46} & \text{33.74} \\
& \text{DP} & \text{35.07} & \text{37.07} & \text{37.51} & \text{37.43} & \text{38.24} & \text{36.91} & \text{36.61} & \text{35.85} & \text{36.51} & \text{33.84} & \text{37.21} & \text{35.74} & \text{34.86} & \text{35.77} & \text{37.86} & \text{36.43} \\
& \text{SP} & \text{38.88} & \text{40.60} & \text{40.21} & \text{40.44} & \text{39.45} & \text{39.37} & \text{40.90} & \text{36.86} & \text{40.61} & \text{37.11} & \text{39.45} & \text{36.26} & \text{35.88} & \text{38.46} & \text{37.35} & \text{38.79} \\
& \text{MP} & \text{32.46} & \text{33.02} & \text{33.98} & \text{32.59} & \text{33.20} & \text{34.54} & \text{34.39} & \text{34.30} & \text{33.90} & \text{33.28} & \text{33.47} & \text{34.69} & \text{32.67} & \text{33.28} & \text{35.68} & \text{33.70} \\
& \text{LTP\_20\%} & \text{40.44} & \text{40.54} & \text{40.86} & \text{41.80} & \text{40.50} & \text{40.66} & \text{41.22} & \text{36.17} & \text{41.04} & \textbf{37.52} & \text{39.34} & \text{35.21} & \text{35.31} & \text{39.18} & \text{37.19} & \text{\underline{39.13}} \\
& \text{LTP\_75\%} & \textbf{41.34} & \textbf{41.30} & \textbf{41.44} & \textbf{42.10} & \textbf{42.81} & \textbf{41.78} & \textbf{42.06} & \textbf{38.80} & \textbf{42.08} & \text{37.43} & \textbf{40.96} & \textbf{39.18} & \textbf{37.43} & \textbf{39.42} & \textbf{41.68} & \textbf{40.65} \\
\hline
\multirow{6}{*}{32} & \text{FT} & \text{35.84} & \text{36.28} & \text{36.00} & \text{36.11} & \text{36.64} & \text{36.02} & \text{36.47} & \text{35.41} & \text{35.68} & \text{35.33} & \text{35.71} & \text{35.90} & \text{34.81} & \text{36.10} & \text{36.20} & \text{35.90} \\
& \text{DP} & \text{41.80} & \text{43.51} & \text{43.49} & \text{42.50} & \text{43.65} & \text{42.83} & \text{43.90} & \text{39.30} & \text{42.39} & \text{37.51} & \text{40.51} & \text{42.01} & \text{39.77} & \text{41.91} & \text{39.94} & \text{\underline{41.67}} \\
& \text{SP} & \text{40.30} & \text{43.38} & \text{42.08} & \text{42.27} & \text{44.72} & \text{42.32} & \text{42.34} & \text{38.91} & \text{43.76} & \text{37.54} & \text{39.97} & \text{38.79} & \text{38.83} & \text{42.09} & \text{39.56} & \text{41.12} \\
& \text{MP} & \text{40.95} & \text{42.16} & \text{42.61} & \text{42.31} & \text{45.52} & \text{41.22} & \text{44.67} & \text{40.17} & \text{42.18} & \text{36.52} & \text{40.16} & \text{41.21} & \text{40.48} & \text{41.74} & \text{40.89} & \text{41.52} \\
& \text{LTP\_20\%} & \text{40.46} & \text{42.83} & \text{41.22} & \text{42.59} & \text{43.65} & \text{42.63} & \text{42.57} & \text{40.42} & \text{43.37} & \text{37.45} & \text{39.32} & \text{39.40} & \text{39.28} & \text{40.72} & \text{41.42} & \text{41.50} \\
& \text{LTP\_75\%} & \textbf{45.33} & \textbf{44.97} & \textbf{43.63} & \textbf{43.99} & \textbf{45.97} & \textbf{43.15} & \textbf{44.93} & \textbf{41.48} & \textbf{44.61} & \textbf{38.32} & \textbf{42.55} & \textbf{42.40} & \textbf{41.12} & \textbf{45.45} & \textbf{43.39} & \textbf{43.42} \\
\hline
\end{tabular}
\end{adjustbox}
\caption{Zero-shot Cross-lingual Transfer performance on XNLI dataset in accuracy (\%). FT, DP, SP, and MP results are taken from \citet{zhao-schutze-2021-discrete}. AVG. is the average of 15 languages. LTP\_20\% and LTP\_75\% denote the framework takes 20\% and 75\% of the parameters of the backbone model separately. Bold font signifies the best results, while underlined font indicates the second-best results.}
\label{tab:xnli}
\end{table*}

\subsection{Implementation details}

For a fair comparison, we utilize XLM-Roberta-base (\citealp{conneau-etal-2020-unsupervised}) as the backbone model, containing 12 layers, 768 hidden units and 12 attention heads. To find the sparse sub-network, we decouple the embedding matrices of the input and output layers, and freeze the output embeddings and LayerNorm layers. We apply the L1 regularization with a coefficient of 0.1. We search the active ratios in [0.5, 0.95] to select the top K relevant parameters in both layer separate and across layers settings. We conduct training for 3 epochs with a batch size of 32, evaluating the validation set, which comprises 10\% of data, every 1000 steps. The initial learning rate is set at 5e-5, and we utilize the AdamW (\citealp{loshchilov2018decoupled}) optimizer. The maximum sequence length is 256. During the prompting and sparse fine-tuning on cross-lingual tasks, we use a batch size of 32 and a learning rate of 2e-5. The prompt length is set to 4 with 256 maximum sequence length. We fine-tune the trainable parameters for 70 epochs and select the best model based on the performance of the development set.

\subsection{Baseline models}

We compare the LTP framework with the work in \citet{zhao-schutze-2021-discrete} on XNLI dataset, which included Fine-tuning, Discrete Prompting, Soft Prompting and Mixed Prompting. For a comprehensive evaluation, we follow the two settings in their work: a) zero-shot cross-lingual transfer: fine-tuning the network using English training data and directly evaluating the performance on the testing data of all languages; b) in-language prompting: fine-tuning the network using a language and evaluating on the same language. We use the same training, development, and testing data in the experiments to conduct few-shot learning. On the AmericasNLI dataset, we compare our approach against Fine-tuning on the Multi-NLI dataset and the Soft Prompting method.

\section{Results and discussion}

In this section, we conduct an empirical investigation into the zero-shot cross-lingual transferability (section~\hyperref[sec:zero-shot]{5.1}) and in-language prompting capabilities (section~\hyperref[sec:in-language]{5.2}) for low-resource languages, both seen and unseen, within pre-trained multilingual language models. Next, we systematically analyze each component of our experimental setting and thus shed light on the design of cross-lingual prompt-learning. 

\subsection{Zero-shot cross-lingual transferability}
\label{sec:zero-shot}

Zero-shot cross-lingual, a standard evaluation setting for natural language inference tasks, enables researchers to assess model transferability without relying on translation tools, unlike the translate-train scenario. We further limit this to a few-shot English dataset to mimic real-world resource scarcity and assess model generalization from minimal data.

\textbf{Seen high-resource and low-resource languages.} We initially illustrate the effectiveness of our proposed LTP framework on high-resource languages and low-resource languages seen in the pre-trained language models. We present the results of the XNLI dataset in Table~\ref{tab:xnli}. The reported results are the average accuracy of 5 runs with different random seeds. LTP framework reaches the best performance with 75\% of the parameters in the original model, and we can see that our framework consistently outperforms baseline methods. Additionally, we present the performance achieved by only using 20\% of parameters, surpassing the baselines by a clear margin in 5 out of 6 distinct experimental settings. The relationship between the number of parameters and performance involves a trade-off, which will be analyzed in the section~\hyperref[sec:analysis]{5.3}. Moreover, a greater improvement is achieved when employing a reduced number of training samples. This confirms that the LTP framework is beneficial for extremely small training sets, even with a substantial parameter reduction. It is worth noting that the LTP framework appears beneficial to low-resource languages like Swahili and Urdu. For example, in the 2-shot setting, the LTP framework with 20\% of parameters enhances Swahili performance from 34.35\% to 36.49\% and improves Urdu from 35.52\% to 36.99\%. LTP framework leads to notable performance improvements for these languages in a parameter-efficient manner, and the benefits can be optimized when we raise the active ratio to 75\%. Meanwhile, the LPT framework consistently delivers improvements across languages originating from distinct language families like Thai, Vietnamese, and Chinese. This further underscores its effectiveness.

\begin{table*}[!ht]
\centering
\begin{adjustbox}{max width=0.95\textwidth}
\begin{tabular}{c|cc|cccccccccc|c}
\hline
\textbf{Method} & \textbf{Source} & \textbf{Size} & \textbf{aym} & \textbf{bzd} & \textbf{cni} & \textbf{gn} & \textbf{hch} & \textbf{nah} & \textbf{oto} & \textbf{quy} & \textbf{shp} & \textbf{tar} & \textbf{AVG.}\\
\hline
\text{FT*} & \text{MultiNLI} & \text{390K} & \text{36.13} & \text{39.65} & \text{37.91} & \text{39.47} & \text{37.20} & \text{42.59} & \text{37.79} & \text{37.24} & \text{40.45} & \text{36.36} & \text{38.48} \\
\hline
\text{SP} & \text{MultiNLI} & \text{32} & \text{35.07} & \text{33.47} & \text{36.67} & \text{36.00} & \text{35.73} & \text{35.37} & \text{35.43} & \text{36.53} & \text{35.87} & \text{35.33} & \text{35.55} \\
\text{SP} & \text{MultiNLI} & \text{64} & \text{37.20} & \text{34.53} & \text{37.47} & \text{37.07} & \text{36.13} & \text{37.13} & \text{37.17} & \text{36.80} & \text{37.33} & \text{38.00} & \text{36.88} \\
\text{LPT\_75\%} & \text{MultiNLI} & \text{32} & \text{37.73} & \text{37.20} & \text{36.80} & \text{38.93} & \text{35.47} & \text{36.72} & \text{36.76} & \text{36.93} & \text{37.20} & \text{36.67} & \text{37.04} \\
\text{LPT\_75\%} & \text{MultiNLI} & \text{64} & \text{41.73} & \text{35.07} & \text{42.67} & \text{40.67} & \text{39.07} & \text{42.28} & \text{39.71} & \text{40.80} & \text{41.87} & \text{40.03} & \textbf{40.40} \\
\hline
\text{SP} & \text{AmericasNLI-dev} & \text{32} & \text{39.73} & \text{41.73} & \text{38.80} & \text{36.67} & \text{36.53} & \text{37.94} & \text{37.97} & \text{39.73} & \text{38.67} & \text{35.87} & \text{38.36} \\
\text{LPT\_20\%} & \text{AmericasNLI-dev} & \text{32} & \text{38.80} & \text{38.67} & \text{40.00} & \text{36.80} & \text{37.47} & \text{38.75} & \text{37.43} & \text{39.20} & \text{38.53} & \text{35.20} & \text{38.09} \\
\text{LPT\_75\%} & \text{AmericasNLI-dev} & \text{32} & \text{42.93} & \text{41.47} & \text{41.60} & \text{37.73} & \text{38.27} & \text{40.11} & \text{41.31} & \text{40.00} & \text{40.93} & \text{37.20} & \text{\underline{40.16}} \\
\hline
\end{tabular}
\end{adjustbox}
\caption{LTP performance on AmericasNLI dataset in accuracy (\%). * is taken from \citet{ebrahimi-etal-2022-americasnli}. AVG. is the average of 10 languages. Source and Size indicate the training set. Bold font signifies the best results, while underlined font indicates the second-best results.}
\label{tab:anli}
\end{table*}

\textbf{Unseen truly low-resource languages.} We further delve into the zero-shot cross-lingual transfer potential for truly low-resource languages that remain unseen by the language models. We report the results of the AmericasNLI in Table~\ref{tab:anli}. We take the results from \citet{ebrahimi-etal-2022-americasnli}, where the XLM-R-base was fine-tuned on 390K English NLI data. We aim to boost performance by leveraging minimal training data, limiting our approach to a few-shot setting. We observe that a 32-shot English dataset proves insufficient for truly low-resource languages, but performance can be improved by utilizing a 64-shot setup. Moreover, reducing trainable parameters to 20\% lowers performance, as the pre-trained model needs more general knowledge of unseen languages. However, the LTP framework with a 75\% active ratio enhances performance by addressing this need. The results further emphasize our LTP framework's robust effectiveness and efficiency.

\begin{table}[!ht]
\centering
\begin{adjustbox}{max width=0.45\textwidth}
\begin{tabular}{c|c|ccccc}
\hline
\textbf{Shots} & \textbf{Method} & \textbf{HI} & \textbf{SW} & \textbf{TR} & \textbf{UR} & \textbf{ZH} \\
\hline
\multirow{2}{*}{1} & \text{SP} & \textbf{35.15} & \text{33.63} & \textbf{34.61} & \text{34.09} & \text{34.53} \\
& \text{LTP\_20\%} & \text{34.61} & \textbf{35.95} & \text{34.33} & \textbf{35.01} & \textbf{35.67} \\
\hline
\multirow{2}{*}{2} & \text{SP} & \text{34.31} & \text{36.05} & \textbf{35.29} & \text{34.33} & \text{34.93} \\
& \text{LTP\_20\%} & \textbf{34.79} & \textbf{36.15} & \text{34.45} & \textbf{34.81} & \textbf{35.47} \\
\hline
\multirow{2}{*}{4} & \text{SP} & \textbf{35.23} & \text{34.21} & \text{39.56} & \textbf{35.21} & \text{34.33} \\
& \text{LTP\_20\%} & \text{34.51} & \textbf{34.35} & \textbf{39.96} & \text{35.11} & \textbf{36.33} \\
\hline
\multirow{2}{*}{8} & \text{SP} & \text{33.93} & \text{33.47} & \text{35.41} & \text{34.59} & \text{34.39} \\
& \text{LTP\_20\%} & \textbf{34.31} & \textbf{33.87} & \textbf{38.10} & \textbf{35.63} & \textbf{34.73} \\
\hline
\multirow{2}{*}{16} & \text{SP} & \text{35.83} & \text{33.99} & \text{40.22} & \text{35.47} & \text{35.64} \\
& \text{LTP\_20\%} & \textbf{38.12} & \textbf{34.77} & \textbf{41.04} & \textbf{38.20} & \textbf{37.49} \\
\hline
\multirow{2}{*}{32} & \text{SP} & \textbf{41.42} & \text{36.00} & \text{43.62} & \text{39.18} & \text{35.13} \\
& \text{LTP\_20\%} & \text{40.58} & \textbf{37.25} & \textbf{43.95} & \textbf{39.64} & \textbf{36.23} \\
\hline
\end{tabular}
\end{adjustbox}
\caption{In-language Prompting performance on XNLI dataset in accuracy (\%). SP results of the 8, 16, and 32-shot settings for Swahili, Turkish, Urdu, and Chinese are taken from \citet{zhao-schutze-2021-discrete}, while the remaining results in SP are re-implemented using the same method as they described.}
\label{tab:in-language}
\vspace{-4mm}
\end{table}

\subsection{In-language prompting}
\label{sec:in-language}

The results above demonstrate the LTP framework's success in capturing and learning knowledge that could be transferred to other languages. However, whether the selected subset of parameters could directly provide information on a particular language remains unexplored. We investigate this question by conducting in-language prompting experiments. The distributions of training and testing sets remain consistent in this case.

\textbf{Seen high-resource and low-resource languages.} We choose Hindi, Turkish, Urdu, Swahili, and Chinese from the XNLI dataset as examples, as they either belong to low-resource categories or represent distinct language families, often resulting in less favorable performance. The results are reported in Table~\ref{tab:in-language}. We observe that training only 20\% of the parameters can lead to improvements compared to the original soft-prompting method. Selecting the English winning ticket proves advantageous for enhancing target languages in the in-language prompting scenario, as English-related parameters contain valuable semantic information. Fine-tuning only these parameters, without significantly altering language-specific knowledge, can achieve some gains with limited computational resources. Additionally, a few results show a minor decrease but remain competitive with zero-shot cross-lingual results. We conjecture that training in English can optimize the utility of those winning tickets, while machine-translated data from the XNLI dataset may not yield as much valuable information for adapting to a specific language as the original English data.

\textbf{Unseen truly low-resource languages.} Even for truly low-resource languages, obtaining a few annotated samples per class is feasible in real-world settings, making evaluating in-language prompting capabilities for these languages meaningful. The AmericasNLI dataset, for example, includes a development set with approximately 750 samples per language. Table~\ref{tab:anli} shows the LTP framework outperforms 390K Multi-NLI fine-tuning with just 32 samples per class and remains competitive with fine-tuning even at 20\% active parameters. Furthermore, it exceeds the performance of Soft Prompting by a clear margin. We conclude that by utilizing a few in-language samples, the LTP framework effectively boosts performance in truly low-resource languages.

\begin{figure*}
\centering
\begin{subfigure}{.5\textwidth}
  \centering
  \includegraphics[width=\linewidth]{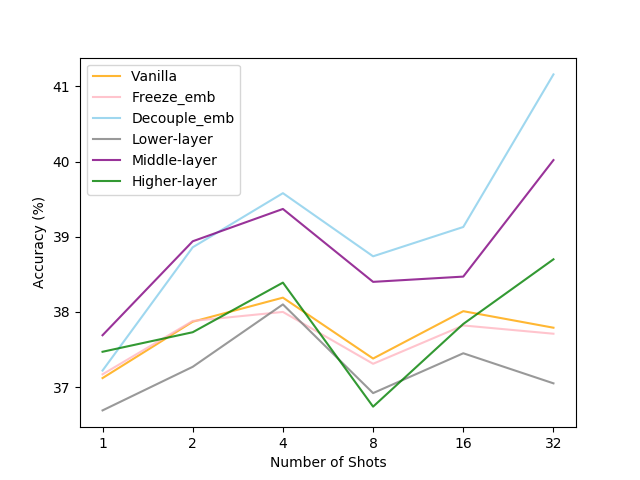}
  \caption{Performance of different strategies and layers.}
  \label{figure:param_select}
\end{subfigure}%
\begin{subfigure}{.5\textwidth}
  \centering
  \includegraphics[width=\linewidth]{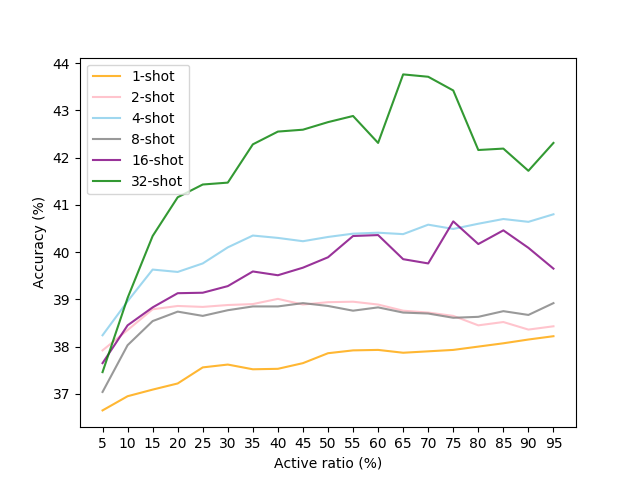}
  \caption{Performance across various active ratios.}
  \label{figure:ratios}
\end{subfigure}
\caption{Detailed analysis of different strategies and active ratios. The performance is the average of 15 languages in the zero-shot cross-lingual transfer setting. In the left figure, we select 20\% as the active ratio.}
\label{figure: test}
\end{figure*}

\subsection{Analysis of parameter selection strategies and active ratios}
\label{sec:analysis}

We systematically conduct a series of ablation studies and thorough analyses to gain deeper insights into the efficacy of different selection strategies and investigate the impact of active ratios. We present the distributions of the active parameters in~\hyperref[sec:fig_dist]{A.1}. Most active parameters are derived from the embedding layer in the vanilla setting. Decoupling the input and output embedding matrices can alleviate this issue. There is a reduced concentration (from 69\% to 39\%) of selected parameters in the embedding matrix compared to the original configuration. The implementation of freezing the embedding layer is incorporated to facilitate a comparative assessment. The performance of different strategies is shown in Figure~\ref{figure:param_select}. We can see that decoupling the input and output matrices outperforms other strategies, since more parameters from the Transformers layers are involved in prompt-learning. These parameters embed more linguistic and semantic knowledge and are beneficial for transferring to other languages. Next, we analyze the trade-off between the number of trainable parameters and performance. We report performance results for various active ratios in Figure~\ref{figure:ratios}. It is straightforward that performance positively correlates with the number of trainable parameters. We find a significant increase in the curve before reaching an active ratio of 20\%, followed by some declines after 75\%. This highlights that employing an excessively low active ratio hinders the acquisition of cross-lingual information, whereas an overly high active ratio disrupts too much language-specific knowledge.

\subsection{Analysis of parameter selection in different layers}

The prevailing trend leans toward scaling up the model size, making it more robust and effectively adapt to various tasks. Reducing the number of trainable parameters in our framework may offer the opportunity for enhanced generalization within a larger model context. We conduct supplementary experiments to provide some insights into further reducing parameters. Previous works (\citealp{jawahar-etal-2019-bert}; \citealp{dalvi2022discovering}) found that lower layers in the Transformers-based models are intricately tied to linguistic information, whereas the higher layers encapsulate semantic knowledge. Moreover, some works (\citealp{de-cao-etal-2020-decisions}; \citealp{su-etal-2022-transferability}; \citealp{xie-etal-2021-importance}) pointed out that the higher layers are more task-specific and language-specific. \citet{hu2020xtreme} conducted extensive experiments on sentence retrieval tasks and noted that sentence representations from the middle layers outperformed those from both the bottom and top layers. Subsequent works (\citealp{chi-etal-2021-infoxlm}; \citealp{chi-etal-2021-improving}; \citealp{tien-steinert-threlkeld-2022-bilingual}) confirmed the robust performance of middle-layer sentence representations and the better alignment across languages.

Inspired by these findings, we hypothesize that certain layers may possess more expressive parameters than others, and it is feasible to achieve comparable results by exclusively extracting the active parameters (the winning tickets) from those layers. In particular, we partition the model into three segments, designating layers 1-4 as lower layers, 5-8 as middle layers, and 9-12 as higher layers. Instead of selecting parameters from all the trainable parameters, we select parameters only from one segment at a time. Note that we do not freeze other segments or the embedding matrix during the parameter selection step. Following the setting mentioned earlier, we untie the output embedding matrix and freeze the LayerNorm and output embeddings. The results of accuracy are presented in Figure~\ref{figure:param_select}. We can observe that the performance of the middle layers is better than the higher and lower layers, which is consistent with prior works. The parameters of middle layers share commonalities across languages (\citealp{10.1145/3561970}) and thus can effectively be tuned exclusively. The performance of the middle layers is slightly better in 1- and 2-shot scenarios compared to selecting parameters across all layers, while the overall number of trainable parameters decreases. From 4 to 32 shots, the performance gap widens as the training samples increase; nevertheless, it still notably outperforms other methods. This presents the opportunity for further parameter reduction while maintaining the performance.

\section{Related works}


To further minimize storage and memory usage, a distinct line of research (\citealp{li-liang-2021-prefix}; \citealp{lester-etal-2021-power}) emerged known as "Fixed-LM Prompt Tuning". They concluded that Prompt-Tuning gained increased competitiveness with scale. \citet{liu-etal-2022-p} demonstrated that well-optimized deep prompt tuning, utilizing continuous prompts for each layer, can consistently achieve performance comparable to fine-tuning across a range of model scales and natural language understanding tasks.

Recently, researchers have delved into exploring prompt-learning in multilingual scenarios, with a predominant focus on cross-lingual inference tasks in most instances. \citet{zhao-schutze-2021-discrete} conducted a comparison between prompt-learning and fine-tuning, showcasing prompt-learning's superior performance in the few-shot cross-lingual inference task. \citet{qi2022enhancing} presented a framework for discrete prompt learning, where an augmented sample is generated by randomly choosing a template in another language. The objective is to minimize the difference between the corresponding answer distributions. \citet{li-etal-2023-enhancing-cross} used bilingual dictionaries to replace some words to generate the augmented samples and aligned answer distributions using a multilingual verbalizer. \citet{qi2022enhancing} and \citet{li-etal-2023-enhancing-cross} leveraged external knowledge from Google Translator or bidirectional dictionaries, which provided additional benefits.

Lottery Ticket Hypothesis (LTH) (\citealp{frankle2018the}) was originally proposed to enhance neural network pruning techniques and has become a highly promising approach. Subsequent studies have introduced different variants and applications of LHT. \citet{Renda2020Comparing} combined LTH with learning rate rewinding. \citet{xia2021robust} applied LTH to reduce the side effects of noisy labels. \citet{ansell-etal-2022-composable} applied LTH to select task-specific and language-specific parameters to improve adaption efficiency. \citet{cunha2022proving} extended LTH to convolutional neural networks.



\section{Conclusion}

In this work, we propose a Lottery Ticket Prompt-learning framework to prompt small-sized language models for cross-lingual tasks targeting on low-resource languages. The proposed framework selects a subset of parameters in the backbone model and fine-tunes them together with prompt-related parameters on the downstream task. This approach minimizes significant alterations to other languages' knowledge when adapting to the source language, thereby enabling effective zero-shot transferability. Extensive experimental results show that LTP outperforms baseline methods on zero-shot cross-lingual transfer and in-language prompting with only 20\% of the parameters. We provide an in-depth analysis of the selection of active ratios and find that 20\% of the parameters can surpass the baseline, while 75\% of the parameters can achieve optimal performance. Meanwhile, we study the different selection strategies and conclude that diverging the concentration of selected parameters from the embedding matrix could contribute to enhancing performance. Furthermore, exclusively choosing active parameters from the middle layers produces results comparable to those from the entire model. This, in turn, allows for a more significant reduction in the size of trainable parameters.  



\bibliography{colm2024_conference}
\bibliographystyle{colm2024_conference}

\appendix
\section{Appendix}

\subsection{Parameter distributions}
\label{sec:fig_dist}
\begin{figure}[h]
  \centering
    \includegraphics[width=8cm]{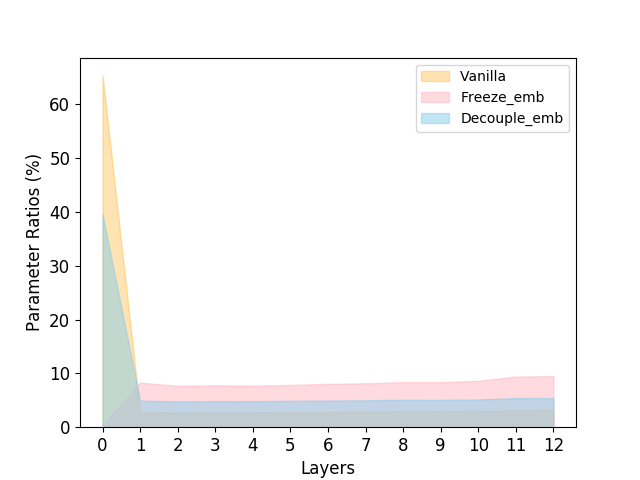}
  \caption{Parameter distributions of different strategies. 0 denotes the embedding layer, while 1-12 corresponds to Transformer layers 1-12.}\label{figure:param_dist}
\end{figure}

\end{document}